\pdfoutput=1
\documentclass[letterpaper, 10 pt, conference]{ieeeconf}  

\IEEEoverridecommandlockouts                              

\usepackage{amsmath} 
\usepackage{amssymb}  
\usepackage{svg}
\usepackage{caption}
\usepackage{multirow}

\title{\LARGE \bf
Fast and Lightweight Scene Regressor for Camera Relocalization
}

\author{Thuan B. Bui$^{1}$, Dinh-Tuan Tran$^{2}$, and Joo-Ho Lee$^{2}$
\thanks{$^{1}$Graduate School of Information Science and Engineering, Ritsumeikan University, Japan
        {\tt\small thuan.aislab@gmail.com}}%
\thanks{$^{2}$College of Information Science and Engineering, Ritsumeikan University, Japan.}%
}

\begin{document}

\maketitle
\thispagestyle{empty}
\pagestyle{empty}

\begin{abstract}

Camera relocalization involving a prior 3D reconstruction plays a crucial role in many mixed reality and robotics applications. Estimating the camera pose directly with respect to pre-built 3D models can be prohibitively expensive for several applications with limited storage and/or communication bandwidth. Although recent scene and absolute pose regression methods have become popular for efficient camera localization, most of them are computation-resource intensive and difficult to obtain a real-time inference with high accuracy constraints. This study proposes a simple scene regression method that requires only a multi-layer perceptron network for mapping scene coordinates to achieve accurate camera pose estimations. The proposed approach uses sparse descriptors to regress the scene coordinates, instead of a dense RGB image. The use of sparse features provides several advantages. First, the proposed regressor network is substantially smaller than those reported in previous studies. This makes our system highly efficient and scalable. Second, the pre-built 3D models provide the most reliable and robust 2D-3D matches. Therefore, learning from them can lead to an awareness of equivalent features and substantially improve the generalization performance. A detailed analysis of our approach and extensive evaluations using existing datasets are provided to support the proposed method. The implementation detail is available at https://github.com/ais-lab/feat2map.

\end{abstract}

\section{INTRODUCTION}
\label{section1}
Precise camera relocalization using visual sensors is a crucial step for many applications including robotics \cite{davison2007monoslam}, autonomous navigation \cite{royer2007monocular}, and augmented reality \cite{castle2008video}. Classical visual localization approaches build a three-dimensional (3D) map, which is constructed based on dense or sparse image pixels, of given environments beforehand. Each cloud point in a 3D sparse cloud map is associated with one or many image descriptors, which are subsequently used to match against two-dimensional (2D) keypoints detected from query images \cite{sattler2016efficient}. Once the 2D-3D matches are established, a robust perspective-n-points (PnP) solver within the random sample consensus (RANSAC) algorithm \cite{sattler2016efficient} is implemented to estimate the camera pose. Although 3D classical-based solutions have demonstrated a favorable recall performance, these solutions cause several issues. For instance, extensive memory and computational power are required when a large 3D map is provided. Additionally, accurate 3D structures from motion (SfM) models are not widely available for consumer-grade devices owing to limited computational resources.  

To obtain solutions with increased efficiency, several approaches have been introduced to learn to surrogate the entire map using only a neural network or to compress the map. In general,  these solutions can be categorized into three classes: (1) The first class is absolute pose regression (APR). The goal of this class method is to learn the entire pipeline of the relocalization system in an end-to-end manner. Although this class demonstrates a potential solution in terms of computing resources,  its localization accuracy is poor \cite{sattler2019understanding}. (2) The second class is scene compression, which aims to reduce memory consumption by removing non-robust and unnecessary cloud points \cite{li2010location} or compressing the descriptors of the 3D points \cite{yang2022scenesqueezer} using quantization. (3) The third common approach is scene coordinate regression, which was first introduced by Shotton et al. \cite{shotton2013scene}. This approach encodes entire 3D scene models using network weights as regressor modules without explicitly storing the 3D models \cite{brachmann2018learning, zhou2020kfnet}. Overall, 3D scene coordinate regression-based methods can achieve state-of-the-art accuracy. However, most of them are computational-resource intensive and reaching real-time inferences is difficult.

In this study, we propose a simple scene regression method that can solve the aforementioned problems while achieving state-of-the-art localization performance. Rather than learning the PnP or RANSAC in a  differentiable manner \cite{brachmann2017dsac, brachmann2019expert}, we keep geometry pose estimation to classic algorithms. Additionally, we do not exploit deep learning to encode entire RGB data into a dense 3D scene model. Instead, we begin our regressor from sparse descriptors. This approach has several advantages: 
\begin{itemize}
    \item \textbf{Lightweight regressor architecture}. Sparse data requires fewer computational resources than that of dense data. The scene regressor network for this work is, therefore, simple and lightweight.
    \item \textbf{Reasoning underlying 2D-3D matches}. Mapping from sparse descriptors to world coordinates can make the regressor aware of the robust equivalent image features. Notably, these matches are extensively corrected beforehand by SfM triangulation and feature matching. Therefore, training a network to reason underlying matched descriptors can lead to a significant enhancement in generalization performance.  
    \item \textbf{Learning reliable descriptors}.  A 3D model created by an SfM method will release reliable descriptors for learning its coordinates. Since these descriptors have passed through a verification step and are matched from different views through SfM, they are most likely reliable and belong to the discriminatory areas in the image.
    \item \textbf{Handling unlabeled data}. Interestingly, choosing to learn the scene coordinates from descriptors presents a new potential to handle any unlabeled data. We can match the key points detected in unlabeled images to the available training descriptors and subsequently train the model in a semi-supervised manner. This provides a straightforward technique to tackle new unlabeled data. However, this study does not exploit this advantage in the experiments. Instead, we demonstrate a simple learning procedure, which is sufficient to achieve state-of-the-art performance. 
\end{itemize}

\begin{figure}[t]
  \centering
  \hspace*{1cm}
  \includegraphics[width=200pt]{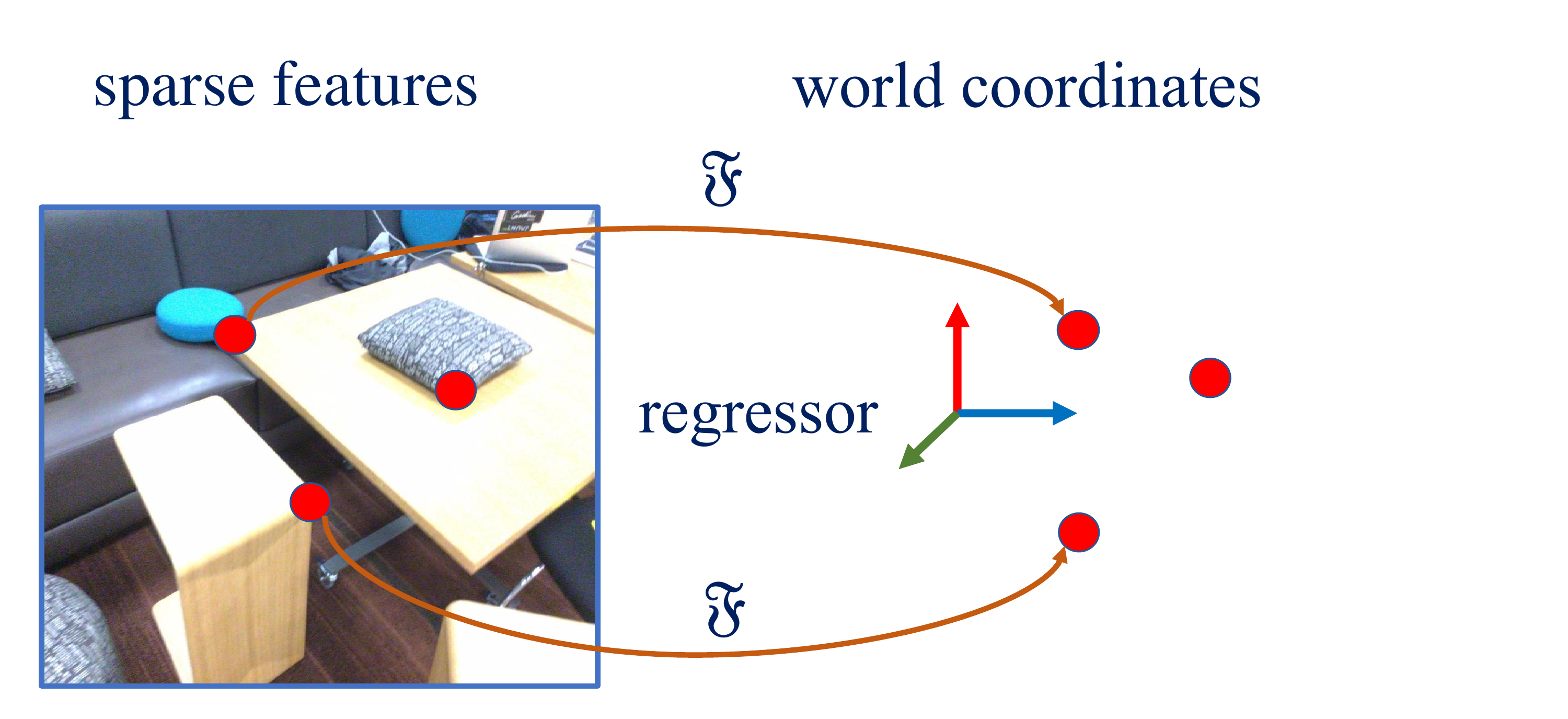}
  \caption{Motivation for this study, in which the sparse map is represented by a simple scene regressor function $\mathfrak{F}$. }
  \label{fig1_motivation}
\end{figure}

The objective of this study is to address camera relocalization problems by satisfying the constraints of low consumption of computational resources and high accuracy. Fig. \ref{fig1_motivation} illustrates the motivation for applying sparse scene regression in this study. Our contributions can be summarized as follows:
\begin{itemize}		
    \item We present a new camera relocalization pipeline and demonstrate that only a simple multi-layer perceptron is sufficient to regress the scene coordinates.
    \item We demonstrate that the proposed regression model is far lightweight and faster than those of recent state-of-the-art dense regression methods and can maintain the same localization errors. 
    \item We conduct an extensive evaluation on two datasets namely, 7scenes and 12scenes, and achieve state-of-the-art performance.
    \item Interestingly, our proposed pipeline can work well by using a tiny regressor network of only  0.7 million parameters and requires only approximately 10\% of the training data to generalize beyond the environments. 
\end{itemize}

\section{RELATED WORK}
Herein, we discuss the studies related to efficient camera relocalization that leverages deep learning.
\subsection{Absolute Pose Regression}
The absolute pose regression approach typically requires a large classifier convolutional neural network that has been pre-trained on a large classification image dataset. This neural network is then fine-tuned to regress the 6DoF pose from either a single \cite{kendall2015posenet, kendall2017geometric} or sequence of images \cite{clark2017vidloc, valada2018deep}. The common practice of this area is first introduced by PoseNet \cite{kendall2015posenet}. Different learning strategies have different constraints either in their
network architectures \cite{walch2017image,brachmann2018learning,wang2020atloc} or loss functions \cite{kendall2017geometric,brachmann2018learning}. To improve network architectures, long short-term memory (LSTM) PoseNet \cite{walch2017image} includes a combination of LSTM and a convolutional neural network to structure latent features. A further improvement in the architecture is achieved by applying the attention mechanism \cite{wang2020atloc} for auto-learning the localization regions of interest. By contrast, the constraints of improving the loss functions are based on the properties of sequence images \cite{brachmann2018learning, valada2018deep} or underlying  3D geometry, such as leveraging reprojection errors \cite{kendall2017geometric}. Owing to the directness of the regression procedure, the execution time is short. However, these efforts do not guarantee that these APR methods can be generalized beyond their training data \cite{sattler2019understanding}. Recent studies additionally leveraged either 3D SfM models \cite{purkait2017spp,bach2022featloc} or Nerf \cite{mildenhall2020nerf,moreau2022lens,chen2022dfnet} to augment more training data. Although these approaches seem promising, they are highly dependent on the quality of an expensive view synthesizer and demonstrate no understanding of the underlying geometric concepts.

\subsection{Scene Compression}
To reduce the number of 3D scene points while maintaining the image-based localization accuracy, Li et al. \cite{li2010location} proposed a K-cover-inspired algorithm to identify a subset of points that sees at least K most quality points. Several subsequent studies have extended this idea either to maximize the visual distinctiveness \cite{cao2014minimal} or visibility \cite{cheng2016data} of the selected K-points. Camposeco et al. \cite{camposeco2019hybrid} proposed a K-cover hybrid-based algorithm to compress two sets of points at different descriptor resolutions. 
Another common scene compression algorithm uses quadratic programming to depict the problem as a convex optimization formulation \cite{mera2020efficient}. This algorithm, therefore, dramatically improves the optimization speed of scene compression. Owing to heavy handcraft and manual hyperparameter tuning of the above approaches, an efficient learning-based method \cite{yang2022scenesqueezer} has been proposed to select the compression points. These approaches are excellent choices for efficient camera relocalization tasks that are especially useful for mobile agents that have limited storage. However, this study focuses on a special class of scene compression, in which an efficient scene coordinate regression is favorable.


\subsection{Scene Coordinate Regression}
The scene coordinate regression approach was proposed by Shotton et al. \cite{shotton2013scene}, in which 2D-3D correspondences are formulated as continuous regression problems. In this approach, estimating accurate camera poses using predicted 3D scene coordinates is straightforward because this replaces expensive matching components with a random forest regressor. Several follow-up studies have improved this approach in terms of accuracy \cite{valentin2015exploiting} and learning camera localization on-the-fly \cite{cavallari2017fly}. Recently, dense coordinates regression that either learns from a single RGB image combined with differentiable PnP RANSAC \cite{brachmann2018learning,brachmann2019expert} or leverages sequential images as a Kalman Filter \cite{zhou2020kfnet}, have become popular. In contrast to these methods, we consider only discriminatory and reliable image regions to regress the scene coordinates. This allows our study to achieve a fast and lightweight localization pipeline while maintaining high localization accuracy. 

\begin{figure}[t]
  \centering
  \includegraphics[width=120pt]{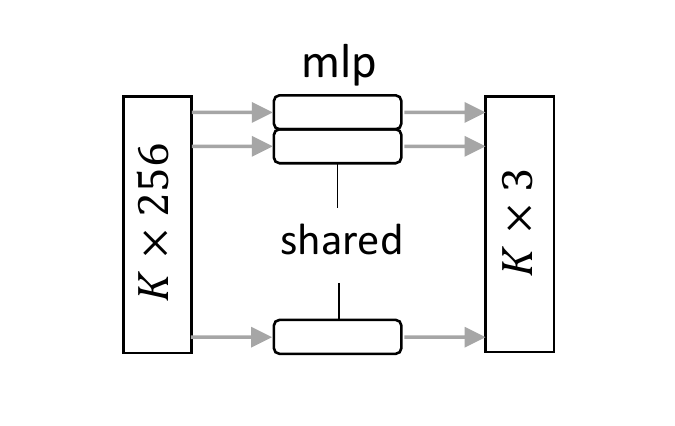}
  \caption{Simple regressor architecture.}
  \label{Architecture}
\end{figure}

\section{PROPOSED METHOD}
In this section, we present the proposed approach for descriptor coordinates regression. Our study is inspired by recent studies on camera relocalization, including absolute pose and scene coordinate regression. The ultimate objectives of these methods are low memory requirement, as well as fast and accurate camera pose estimation within an inference time. To achieve these objectives, we propose a straightforward method that requires lower computational power than those required by the above methods. In the following sections, we present the details of the proposed method for scene regression, including the loss function used to train the network.

\begin{table*}[h]
\centering
\caption{The median position and rotation errors of different relocalization methods on \textit{7scenes} dataset. Best results are
in bold.}
\begin{tabular}{c|c|c|c|c|c|c|c|c}
\hline
\hline
  \multirow{2}{*}{Scene} &\multirow{2}{*}{Volume}& E-PoseNet &  FeatLoc  & CamNet & Active  & DSAC++ & SCoordNet & \multirow{2}{*}{Proposed} \\
   & & \cite{musallam2022leveraging} &  \cite{bach2022featloc}  &  \cite{ding2019camnet}& Search \cite{sattler2016efficient} & \cite{brachmann2018learning}& \cite{zhou2020kfnet} &\\
\hline
  Chess      & 6m$^{3}$  &0.08m, 2.57$^\circ$ & 0.07m, 3.66$^\circ$   &0.04m, 1.73$^\circ$    & 0.04m, 1.96$^\circ$  & 0.02m, 0.5$^\circ$ &  0.02m,  0.63$^\circ$ &  0.02m, 0.80$^\circ$\\
 
  Fire       & 2.5m$^{3}$  & 0.21m, 11.0$^\circ$& 0.17m, 5.95$^\circ$   &  0.03m, 1.74$^\circ$  & 0.03m, 1.53$^\circ$  & 0.02m, 0.9$^\circ$ & 0.02m, 0.91$^\circ$ & 0.03m, 1.13$^\circ$\\
 
  Heads      & 1m$^{3}$   &  0.16m, 10.3$^\circ$& 0.10m, 7.57$^\circ$   &  0.05m, 1.98$^\circ$  & 0.02m, 1.45$^\circ$  & 0.01m, 0.8$^\circ$& 0.02m, 1.26$^\circ$ & 0.01m, 0.90$^\circ$\\
 
  Office     & 7.5m$^{3}$ & 0.15m, 6.80$^\circ$ & 0.16m, 5.20$^\circ$   &  0.04m, 1.62$^\circ$  & 0.09m, 3.61$^\circ$  & 0.03m, 0.7$^\circ$& 0.03m, 0.73$^\circ$ & 0.03m, 0.87$^\circ$\\
 
  Pumpkin    & 5m$^{3}$   & 0.16m, 3.82$^\circ$ & 0.11m, 3.86$^\circ$   &  0.04m, 1.64$^\circ$  & 0.08m, 3.10$^\circ$  & 0.04m, 1.1$^\circ$& 0.04m, 1.09$^\circ$& 0.05m, 1.38$^\circ$\\
 
  RedKitchen & 18m$^{3}$  & 0.20m, 6.81$^\circ$ & 0.20m, 6.43$^\circ$   &  0.04m, 1.63$^\circ$  & 0.07m, 3.37$^\circ$  & 0.04m, 1.1$^\circ$& 0.04m, 1.18$^\circ$ & 0.05m, 1.54$^\circ$\\
 
  Stairs     & 7.5m$^{3}$ & 0.24m, 9.92$^\circ$ & 0.16m, 8.57$^\circ$   &  0.04m, 1.51$^\circ$  & 0.03m, 2.22$^\circ$  & 0.09m, 2.6$^\circ$& 0.04m, 1.06$^\circ$& 0.11m, 3.37$^\circ$\\
 
\hline
  Average &  &0.17m, 7.32$^\circ$  & 0.14m, 5.89$^\circ$& 0.04m, 1.69 $^\circ$& 0.05m, 2.46 $^\circ$& 0.04m, 1.10$^\circ$& \textbf{0.03m, 0.98}$^\circ$ & 0.04m, 1.43$^\circ$\\
\hline

\end{tabular}
\label{all_results}
\end{table*}
\begin{figure*}
  \centering
  \includegraphics[width=110mm]{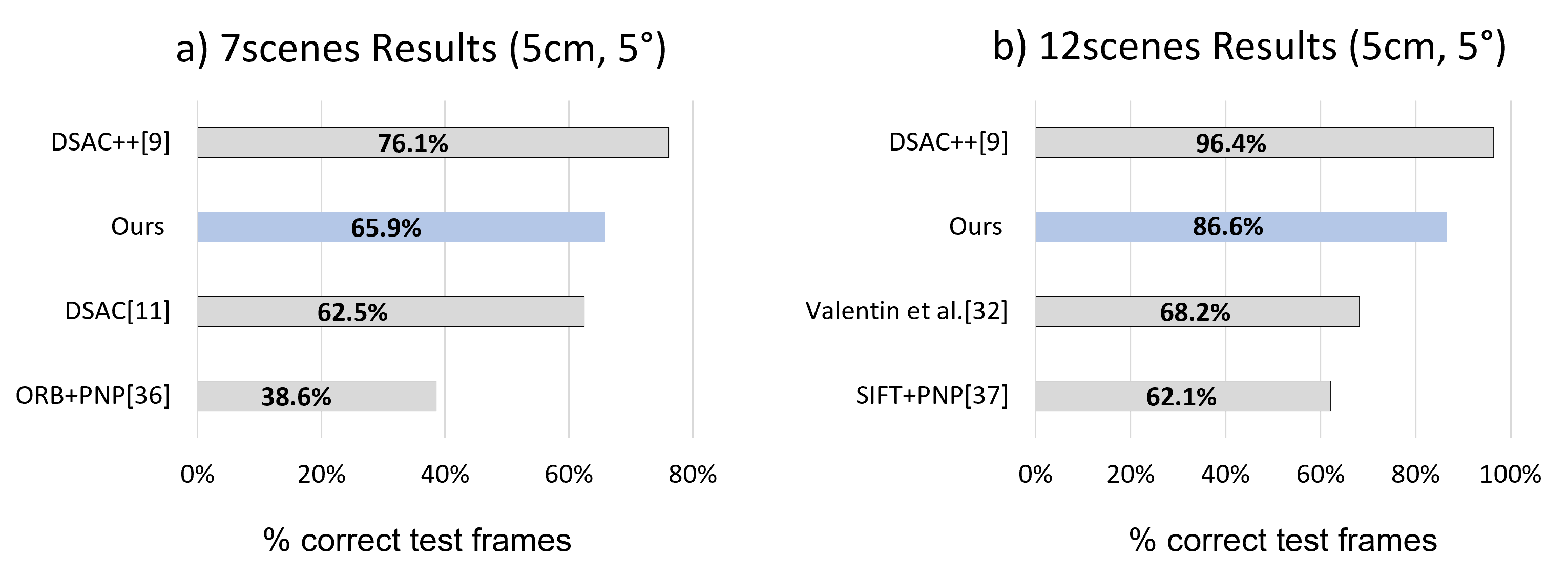}
  \caption{Localization accuracy obtained using the \textit{7scenes} and \textit{12scenes} datasets. The percentages of accurate test frames with pose error below 5 cm, 5$^{\circ}$ obtained using the \textit{7scenes} \textbf{(a)} and 12scenes \textbf{(b)} datasets are presented. The results obtained using the proposed method are marked in blue. }
  \label{Localization_Accuracy}
\end{figure*}

\begin{table}
    \centering
    \caption{The median position and rotation errors of the proposed method against FeatLoc \cite{bach2022featloc} on \textit{12scenes} dataset.}
    \begin{tabular}{c|c|c|c}
\hline
\hline
 \multirow{2}{*}{Scene} &\multirow{2}{*}{Volume} &  FeatLoc &\multirow{2}{*}{Proposed} \\
  &  & \cite{bach2022featloc}  & \\
\hline
A1 Kitchen  & 33m$^{3}$ & 0.32m, 5.19$^\circ$ &  0.02m, 0.83$^\circ$\\
A1 Living   & 30m$^{3}$ & 0.26m, 3.89$^\circ$ &  0.03m, 0.72$^\circ$\\
A2 Bed      & 14m$^{3}$ & 0.37m, 5.39$^\circ$ &  0.02m, 0.99$^\circ$\\
A2 Kitchen  & 21m$^{3}$ & 0.73m, 6.37$^\circ$ &  0.01m, 0.40$^\circ$\\
A2 Living   & 42m$^{3}$ & 0.40m, 5.71$^\circ$ &  0.04m, 1.14$^\circ$\\
A2 Luke     & 53m$^{3}$ & 0.33m, 4.85$^\circ$ &  0.02m, 1.03$^\circ$\\
O1 Gates362 & 29m$^{3}$ & 0.52m, 5.22$^\circ$ & 0.01m, 0.30$^\circ$\\
O1 Gates381 & 44m$^{3}$ & 0.42m, 6.23$^\circ$ &  0.02m, 1.01$^\circ$\\
O1 Lounge   & 38m$^{3}$ & 0.39m, 4.50$^\circ$ &  0.03m, 0.84$^\circ$\\
O1 Manolis  & 50m$^{3}$ & 0.30m, 4.67$^\circ$ &  0.02m, 0.82$^\circ$\\
O2 5a       & 38m$^{3}$ & 0.31m, 4.32$^\circ$ &  0.03m, 1.35$^\circ$\\
O2 5b       & 79m$^{3}$ & 0.23m, 4.14$^\circ$ &  0.03m, 0.94$^\circ$\\
\hline
 Average    &   & 0.38m, 5.04$^\circ$ &  \textbf{0.02m, 0.86}$^\circ$\\
 \hline
    \end{tabular}
    \label{results_12scenes_table}
\end{table}

\subsection{Overview of the Method}
Given a sparse feature set $\mathcal{P}_{t}=\{\mathbf{d}_{i}\in \mathbb{R}^{M}|i=1,...,k\}$ extracted from a RGB image  $\mathcal{I}_{t}$, where $\mathbf{d}$ is a descriptor vector with a dimension of $M$ . In this study, we identify an accurate estimation of the world coordinate $\mathcal{W}_{t}=\{\mathbf{w}_{i}\in \mathbb{R}^{3}|i=1,...,k\}$ by proposing a simple elements set regressor $\mathfrak{F}(\mathcal{P}_{t}|\theta) = \mathcal{W}_{t}$. Here, $\mathfrak{F}$ is typically a deep neural network and $\theta$ is its network weight. As the regressor receives the sparse features as the input, the proposed network is remarkably lightweight and fast for relocalization. The details of the proposed regressor are presented in the next section.

\subsection{Regressor Architecture}
In this section, we propose a simple network architecture for learning to regress the world coordinates of sparse features. The model accepts a descriptors set of $\{\mathbf{d}_{1},...,\mathbf{d}_{k}\}$ as the input and outputs a corresponding world coordinates of $\{\mathbf{w}_{1},...,\mathbf{w}_{k}\}$. Notably, the extracted sparse descriptors set has two properties, i.e., it is an unordered set and has a variable number of descriptors. Therefore, the proposed architecture must be a function of the elements set as follows: 

\begin{equation}
\begin{aligned}
\mathfrak{F}(\{\mathbf{d}_{1},...,\mathbf{d}_{k}\}) & = \{h(\mathbf{d}_{1}),...,h(\mathbf{d}_{k})\}\\
& = \{\mathbf{w}_{1},...,\mathbf{w}_{k}\},
\end{aligned}
\end{equation}

where $\mathfrak{F}:\mathbb{R}^{K \times M} \to \mathbb{R}^{K \times 3}$ and $h:\mathbb{R}^{M} \to \mathbb{R}^{3}$. 

Empirically, the proposed module is simple. We only use a shared non-linear function $h$ for every input descriptor, which is approximated as a multi-layer perceptron (MLP) network. The final network architecture, therefore, is simply a shared-parallel MLP function. 

In the proposed module, we assume that the extracted descriptors have their own distinctiveness and pair-wise matching proximity. With this assumption, the proposed regressor architecture is well-suited to environments that have a rich texture and less similar regions. Our proposed module is displayed in Fig. \ref{Architecture}, in which the $mlp$ function shares its weights with every input descriptor.

\subsection{Loss Function}
Once the scene coordinates are computed, we minimize the following loss function to train the network:

\begin{equation}
Loss(\theta) = \frac{1}{\lvert N \lvert}\sum\limits_{i}^N\lVert \mathcal{\hat{W}}_{i} - \mathcal{W}_{i} \lVert_{2}
\end{equation}

where  $\mathcal{\hat{W}} = \{\mathcal{\hat{W}}_{i},...,\mathcal{\hat{W}}_{N}\}$ are the predicted coordinates, $\mathcal{W}=\{\mathcal{W}_{i},...,\mathcal{W}_{N}\}$ are the ground truth coordinates, and $N$ is the mini-batch size.

This loss function can be improved using an additional method, which uses a common reprojection loss combined with above function \cite{brachmann2018learning}. However, during the experiments, we identified that this combination is prone to get stuck at the local minimums and is unstable during training. Therefore, we simply used this function to train all subsequent experiments.

\section{EXPERIMENTS}

In this section, we evaluated the proposed relocalization pipeline and compared it with standard benchmarks to test the robustness of the pipeline. We subsequently conducted an extensive ablation study to confirm our hypotheses of choosing to regress scene coordinates from sparse descriptors. 

\subsection{Datasets}
In accordance with previous studies \cite{bach2022featloc, brachmann2018learning, ding2019camnet, zhou2020kfnet}, we used two datasets namely, \textit{7scenes} \cite{shotton2013scene} and \textit{12scenes} \cite{valentin2016learning} to evaluate the proposed method. \textit{7scenes} is a collection of RGB-D sequence images that were captured by a handheld Kinect RGB-D camera and consists of seven different indoor environments, in which the spatial extent is less than 4 m. This dataset poses certain difficulties for most localization pipelines using sparse features, such as texture-less regions in the \textit{stairs} scene or poor-quality blurred images. By contrast, \textit{12scenes} contains images with better quality and larger environments than those of \textit{7scenes}. The training sets of \textit{12scenes} are smaller and contain only several hundred frames, whereas \textit{7scenes} has several thousand images in its training and testing sequences. 
\subsection{Architecture and Setup}
The architecture comprised several conventional MLP layers, and we used SuperPoint \cite{detone2018superpoint} descriptors as the input features. Therefore, the input dimension was $\mathbf{d}_{i}\in \mathbb{R}^{256}$, and the final regressor architecture setup was MLPs $(512,1024,1024,512,3)$. Additionally, ReLU was used as the nonlinearity activator for the entire network, excluding the last layer, for producing scene coordinates. 

To generate the ground truth scene coordinates, we reconstructed the SfM models of the entire \textit{7scenes} and \textit{12scenes} datasets with a default image size of $640 \times 480$. We subsequently used the depth information to refine the constructed 3D cloud maps to make scene coordinate labels reliable for subsequent evaluations. Notably, the final inference pipeline required only RGB images to estimate the scene coordinates. For the camera pose estimation, we used PnP-RANSAC, which was implemented in pycolmap  python binding using the maximum error pixels of 12. 

We implemented our algorithm using Pytorch \cite{ketkar2017introduction} with the ADAM \cite{kingma2014adam} optimizer by considering $\beta_{1}=0.9, \beta_{2} = 0.999$, and a weight decay of $5\times10^{-4}$. We set the initial learning rate to $1e^{-3}$ with a decay of $0.5$ applied at each one-fifth of the total epochs. All experiments were conducted on an NVIDIA GTX 1080ti GPU, in which the regressor network requires approximately 1 ms of run-time. In accordance with previous studies \cite{bach2022featloc, wang2020atloc, brachmann2018learning, zhou2020kfnet, ding2019camnet}, we computed the median errors for position and orientation. All the experiments were trained using 1000 epochs, and the mini-batch size was $8$. 

\subsection{Baselines}
We compared our results with those obtained from recent APR methods that learn from sparse features \cite{bach2022featloc} or dense RGB images \cite{moreau2022lens, musallam2022leveraging}. These studies improved localization performance using a large amount of additional synthetic data; however, the proposed method can use a significant small amount of training data for learning. More importantly, the proposed method exploits the 3D scene geometry for learning and subsequently uses classical algorithms such as PnP and RANSAC to estimate the camera poses. Therefore, we compared the proposed method with Active Search \cite{sattler2016efficient}, which is a long-standing benchmark for geometry-based methods with all the components being classical. We additionally compared our method with that reported by Valentin et al. \cite{valentin2016learning}, DSAC \cite{brachmann2017dsac},  DSAC++\cite{brachmann2018learning}, and SCoordNet \cite{zhou2020kfnet} which also apply scene regression. We further compared our method with the recent state-of-the-art relative pose regression method i.e., CamNet \cite{ding2019camnet}. For the listing of the results of the baselines, we report the results mentioned by their original publications, with the exception of the results obtained by Active Search, which were reported by  \cite{zhou2020kfnet, brachmann2018learning}.

\begin{table}[h]
\centering
\caption{Accuracy within different thresholds on \textit{7scenes} and \textit{12scenes} datasets.}
\begin{tabular}{lclclclc}
\hline
\hline
Threshold & 7scenes & 12scenes \\
\hline
3cm, 3$^\circ$ & 25\% & 64\% \\
5cm, 5$^\circ$ & 66\% & 87\% \\
10cm, 5$^\circ$ & 85\% & 97\% \\
\hline
\end{tabular}
\label{12scenes_3_thresholds}
\end{table}

\begin{figure*}
  \centering
  \includegraphics[width=150mm]{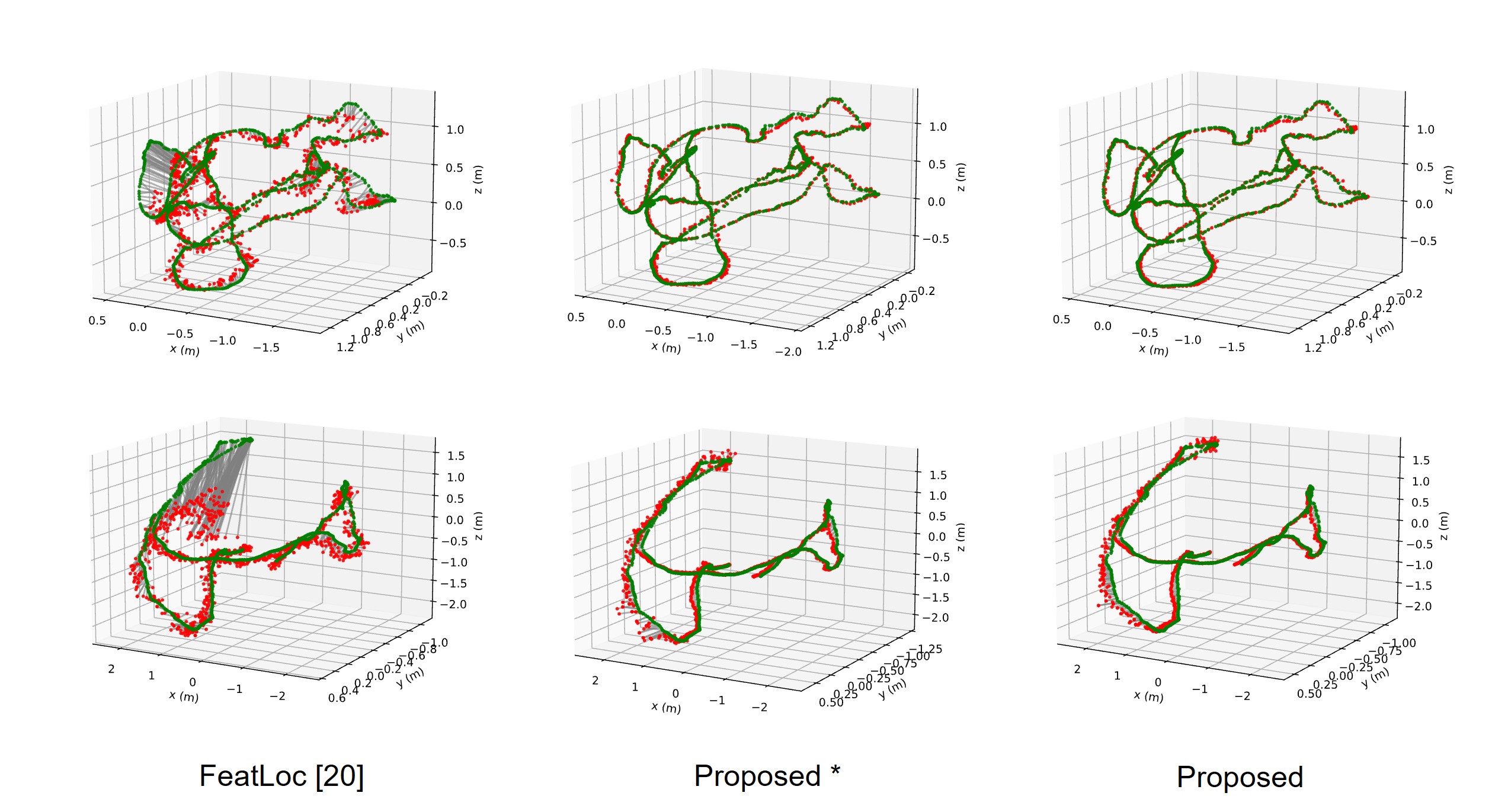}
  \caption{Camera relocalization results on Chess-seq-03 and Redkitchen-seq-12 from the \textit{7scenes} dataset. The green points represent the ground truths and red points represent the predictions. The result marked with a star indicates the pipeline using only a tiny regressor network of 0.7 million parameters.}
  \label{ches_red_visualization}
\end{figure*}

\subsection{Relocalization Performance}
    \textbf{Localization Errors}. Table \ref{all_results} lists the median localization errors such as translation and rotation errors corresponding to the \textit{7scenes} dataset. The proposed approach achieved the same position errors for the \textit{Chess}, \textit{Heads}, and \textit{Office} scenes as those of DSAC++ \cite{brachmann2018learning}, and ScoordNet \cite{zhou2020kfnet}. On the average of the total scenes, the proposed approach approximately obtained the same translation error as that of DSAC++, and, CamNet \cite{ding2019camnet}. However, the rotational error was higher than about $0.33^{\circ}$ and $0.45^{\circ}$ as that of DSAC++ and SCoordNet respectively. In contrast, the proposed method outperformed the conventional baseline of Active Search \cite{sattler2016efficient} and that of absolute pose regression FeatLoc \cite{bach2022featloc}, and E-PoseNet \cite{musallam2022leveraging}. Fig. \ref{Localization_Accuracy}a demonstrates a comparison in terms of classification accuracy of the proposed method with those of ORB + PnP \cite{schindler2007city}, DSAC \cite{brachmann2017dsac}, and DSAC++. The results were calculated based on the threshold of $5 cm, 5^{\circ}$, and our method achieved approximately 66\% accuracy. DSAC++ obtained a better one of 76.1\%.

Similarly, we used the same comparison criteria for the \textit{12scenes} dataset, as seen in the Table. \ref{results_12scenes_table}. However, previous studies have not reported the median localization errors pertaining to this dataset. Therefore, we compared our proposed method with a recent APR method namely, FeatLoc \cite{bach2022featloc}. FeatLoc is a lightweight APR approach that learns from sparse features, similar to our approach. In details, the proposed method exhibited substantially lower translational and rotational errors than those of FeatLoc. In Fig. \ref{Localization_Accuracy}b, we compared the accuracy of our approach with those of the previous 3D scene learning-based approaches. As shown in the figure, DSAC++ \cite{brachmann2018learning} achieves the highest accuracy of 96.4\%, whereas this study reached approximately 86.6\% accuracy. The proposed method outperformed the methods reported by Valentin et al. \cite{valentin2016learning}, and Svarm et al. \cite{svarm2014accurate} with a difference of approximately 18.4\% and 24.5\% respectively. For some applications that do not require this level of precision, our method scored 97\% within $10 cm,5^{\circ}$, as seen in Table \ref{12scenes_3_thresholds}. Fig. \ref{ches_red_visualization} displays some of the camera relocalization results.

\begin{table}[h]
\centering
\caption{Comparison on the number parameters and  running time of different scene regressor networks.}
\begin{tabular}{lclclclc}
\hline
\hline
Method & DSAC++ \cite{brachmann2018learning} & SCoordNet \cite{zhou2020kfnet} & Proposed\\
\hline
\#params  & 210M  & \hfil24M & \textbf{3.6M}\\
Time & 486.07ms & \hfil156.6ms & \textbf{14ms}\\
\hline
\end{tabular}
\label{system_efficent}
\end{table}

\begin{table}[!htbp]
\centering
\caption{Localization accuracy of a significant small regressor network, which has only 0.7M parameters.}
\begin{tabular}{lclclclc}
\hline
  Threshold &   \multicolumn{2}{c}{7scenes} &      &  \multicolumn{2}{c}{12scenes} &      \\
\hline
       &    Accuracy &  Median Errors &  Accuracy &  Median Errors \\
\cline{2-5}
    5cm, 5$^\circ$  &  47.0\% & \multirow{2}{*}{5.3cm, 1.64$^\circ$} & 69.1\% & \multirow{2}{*}{3.7cm, 1.46$^\circ$} \\
    10cm, 5$^\circ$  &   81.5\%  &   & 90.6\%  \\
\hline
\end{tabular}
\label{0_7Msresults}
\end{table}

\begin{table*}
\centering
\caption{Median localization errors of the proposed method when learning with different numbers of training data, against LENS\cite{moreau2022lens}. The results show that the proposed method only requires about 10\% amount of training data to reach the same performance as that of LENS\cite{moreau2022lens}, where LENS requires 1100\% amount of training data. The in bold results are the nearest ones which have an equal or lower error compared with that of LENS.}
\begin{tabular}{c|c|c|c|c|c|c|c}
\hline
\hline
 \multirow{3}{*}{Scene} & \multicolumn{5}{c}{The proposed} & & LENS \cite{moreau2022lens} \\
\cline{2-8}
          & \multirow{2}{*}{100\%} & \multirow{2}{*}{80\%} & \multirow{2}{*}{60\%} & \multirow{2}{*}{40\%} & \multirow{2}{*}{20\%} & \multirow{2}{*}{10\%} & \multirow{2}{*}{$ real. + \frac{syn.}{real.}=1,100\%$}  \\
          & & & & & & & \\
\hline
Chess      & 0.02m, 0.8$^\circ$ & 0.02m, 0.8$^\circ$ & 0.02m, 0.8$^\circ$ & 0.03m, 0.9$^\circ$ & \textbf{0.03m}, 1.0$^\circ$ & 0.04m, \textbf{1.3}$^\circ$ & 0.03m, 1.3$^\circ$\\
Fire       & 0.03m, 1.1$^\circ$ & 0.03m, 1.3$^\circ$ & 0.04m, 1.4$^\circ$ & 0.04m, 1.5$^\circ$ & 0.06m, 2.2$^\circ$ &\textbf{ 0.09m, 3.0}$^\circ$ & 0.10m, 3.7$^\circ$\\
Heads      & 0.01m, 0.9$^\circ$ & 0.02m, 1.2$^\circ$ & 0.02m, 1.2$^\circ$ & 0.03m, 1.7$^\circ$ & 0.04m, 2.3$^\circ$ & \textbf{0.06m, 3.7}$^\circ$ & 0.07m, 5.8$^\circ$\\
Office     & 0.03m, 0.9$^\circ$ & 0.03m, 1.0$^\circ$ & 0.04m, 1.0$^\circ$ & 0.04m, 1.1$^\circ$ & \textbf{0.05m, 1.3}$^\circ$ & 0.08m, 2.0$^\circ$& 0.07m, 1.9$^\circ$\\
Pumpkin    & 0.05m, 1.4$^\circ$ & 0.05m, 1.4$^\circ$ & 0.05m, 1.5$^\circ$ & 0.05m, 1.5$^\circ$ & 0.06m, 1.5$^\circ$ & \textbf{0.08m, 1.8}$^\circ$& 0.08m, 2.2$^\circ$\\
Redkitchen & 0.05m, 1.5$^\circ$ & 0.06m, 1.6$^\circ$ & 0.06m, 1.6$^\circ$ & 0.06m, 1.7$^\circ$ & \textbf{0.08m, 1.9}$^\circ$ & 0.10m, 2.3$^\circ$& 0.09m, 2.2$^\circ$\\
Stairs     & \textbf{0.11m, 3.4}$^\circ$ & 0.18m, 5.1$^\circ$ & 0.20m, 5.6$^\circ$ & 0.23m, 6.1$^\circ$ & 0.31m, 8.1$^\circ$ & 0.57m,13.8$^\circ$& 0.14m, 3.6$^\circ$\\
\hline
\end{tabular}
\label{learningless}
\end{table*}

\textbf{Systems Efficiency}. 
As reported in the previous subsections, our proposed method can outperform those reported in previous studies in terms of median localization errors, while obtaining approximately the same results as those of DSAC++ \cite{brachmann2018learning} and SCoorNet \cite{zhou2020kfnet}. However, the proposed pipeline is cost-effective, unlike the other methods. The scene regressor in our method requires only 2.3 million parameters, whereas DSAC++ requires 210 million parameters. Notably, this study uses SuperPoint \cite{detone2018superpoint} as the feature extractor. The total number of parameters is, therefore, 3.6 million. Owing to the large network architecture, DSAC++ has a runtime of 486 ms for producing the scene coordinates. A single feedforward from the descriptors of the proposed method requires only approximately 1 ms with 2048 input descriptors. In total, from an RGB image with the size of $640\times480$, the proposed pipeline runs within 14 ms, which is $\times34$ faster than that of DSAC++ and $\times11$ faster than that of SCoordNet \cite{zhou2020kfnet}. These comparisons are summarized in Table \ref{system_efficent}.

\textbf{Simple Regressor Achitecture}.
Here, we examine the proposed pipeline using a small regressor network to determine the efficiency of the method when learning from sparse descriptors. The network setting for this experiment is as follows: MLPs$(512,512,512,128,3)$, where ReLU is the activation function of the network. The total number of learnable parameters was only 0.7 million. Table \ref{0_7Msresults} lists the average median localization errors pertaining to the \textit{7scenes} and \textit{12scenes} datasets. Additionally, we report the classification accuracies obtained using these datasets within two different thresholds: $5 cm, 5^{\circ}$ and $10 cm, 5^{\circ}$. With a small network of 0.7 million parameters, our approach achieved classification accuracies of 81.5\% and 90.6\% within $10 cm, 5^{\circ}$ corresponding to the \textit{7scenes} and \textit{12scenes} datasets, respectively. In comparison with previous studies, this result is even superior to that of Active Search \cite{sattler2016efficient} in terms of rotation errors, while presenting its outstanding  lightweight system.


\subsection{Generalization Performance}

This section reports the generalization performance of the proposed method when learning using less training data. We trained the proposed network with a decreasing amount of training data from 100\% to 10\% in a random sample manner. Table. \ref{learningless} reports the median localization errors in comparison with those of LENS \cite{moreau2022lens} at different percentages. With only a random selection of 60\% data, our pipeline can approximately maintain the same localization errors as those obtained using 100\% data. Interestingly, when learning with only 10-20\% of the data, the proposed method can outperform a recent state-of-the-art APR approach of LENS. Notably, this study \cite{moreau2022lens} has been improved by an additional 1000\% synthetic images about the environments to reach this baseline. This point of view has shown an important confirmation of the proposed approach to generalization performance.

\begin{figure}[t]
  \centering
  \includegraphics[width=250pt]{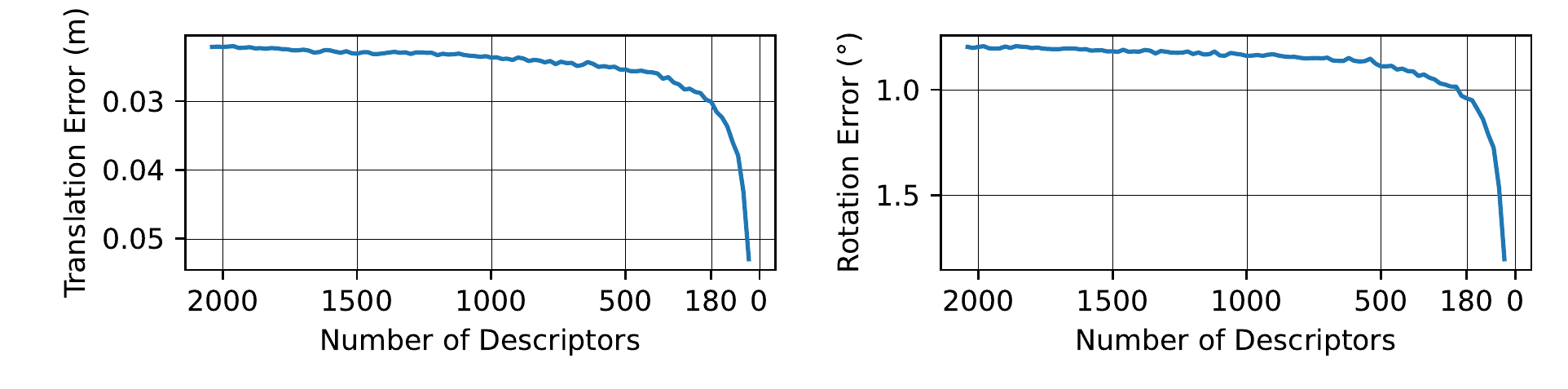}
  \caption{Median localization errors pertaining to the \textit{Chess} scene against the number of input descriptors per image.}
  \label{ches_decreasing}
\end{figure}

Furthermore, we conducted an ablation study on the behavior of the proposed pipeline by reducing the number of input descriptors. Fig. \ref{ches_decreasing} illustrates the changes in the median localization errors, including translational and rotational errors, on the \textit{Chess} scene. All experiments were trained on 2048 descriptors per image. The model was found to be stable over 180 input descriptors. For instance, the pipeline obtained median localization errors of $0.022 m, 0.794^{\circ}$ at 2048 descriptors, $0.024 m, 0.864^{\circ}$ at 640 descriptors, and $0.030 m, 1.04^{\circ}$ at 180 descriptors. Additionally, the pipeline obtained a high error of $0.053 m, 1.80^{\circ}$ at only 40 input descriptors. However, the model can be considered to be generalizable with the change of number input descriptors.

\section{CONCLUSIONS}
In this study, we proposed a simple approach for scene coordinates regression to achieve efficient camera relocalization. The proposed method is simple and achieves state-of-the-art performance. Our method achieved 85\% and 97\% accuracy within $10 cm, 5^{\circ}$ corresponding to the \textit{7scenes} and \textit{12scenes} datasets, respectively, and operated $\times11$  faster than recent state-of-the-art scene regression methods \cite{brachmann2018learning,zhou2020kfnet}. Interestingly, the proposed approach requires only approximately 20\% of the training data to outperform a recent APR method \cite{moreau2022lens}, which has been improved by an expensive data augmentation procedure.   

In future work, we aim to pursue further uses of unlabeled data as a training data source and explore the probabilistic filtering applied to dynamic environments of this algorithm. 

\bibliographystyle{unsrt}
\bibliography{reference.bib}

\end{document}